\documentclass{article}

\usepackage[preprint]{paper_style}

\usepackage[utf8]{inputenc}
\usepackage[T1]{fontenc}
\usepackage{microtype}
\usepackage{graphicx}
\usepackage{subcaption}
\usepackage{booktabs}
\usepackage{longtable}
\usepackage[table]{xcolor}
\usepackage{hyperref}
\usepackage{url}
\usepackage{xurl}
\usepackage{amsmath}
\usepackage{amssymb}
\usepackage{amsfonts}
\usepackage{mathtools}
\usepackage{amsthm}
\usepackage{nicefrac}
\usepackage{multirow}
\usepackage{comment}

\usepackage[capitalize,noabbrev]{cleveref}

\theoremstyle{plain}

\theoremstyle{definition}

\theoremstyle{remark}

\usepackage[disable,textsize=tiny]{todonotes}

\title{When Models Disagree: Rethinking LLM Evaluation for Public Comment Analysis}

\author{%
  Aisha Najera \\
  AI Lab, Princeton University \\
  Engineering and Applied Sciences, RAND Corporation \\
  \texttt{aishanajera@gmail.com} \\
  \And
  Alvin Moon \\
  Engineering and Applied Sciences \\
  RAND Corporation \\
  \texttt{alvinmoon@gmail.com} \\
  \And
  Vedant Srinivasan \\
  Science, Technology, and International Affairs \\
  Georgetown University \\
  \texttt{vs736@georgetown.edu} \\
  \And
  Rajesh Veeraraghavan\thanks{Corresponding author.} \\
  Science, Technology, and International Affairs \\
  Georgetown University \\
  \texttt{rv408@georgetown.edu} \\
}

\begin{document}

\maketitle

\begin{abstract}

Federal agencies are deploying large language models (LLMs) to categorize public comment corpora, where the model's organization of
  the record shapes what policymakers see and which arguments register. Standard evaluation, anchored on stance accuracy against a small
   validated set, cannot detect when different models produce materially different categorizations of the same public input. We propose 
  an Interpretive Audit Pipeline that treats multi-model disagreement as diagnostic of interpretive complexity and directs human review
   toward genuinely ambiguous public input. Analyzing 1,260 public comments on a federal USDA docket across four LLMs, we find that     
  inter-model thematic divergence exceeds within-model prompt variation, and that an expert rubric suppresses deep interpretive
  disagreement without resolving it. In a two-stage labeling study on a stratified 40-comment subsample, four LLMs and a human annotator
   labeled independently and then revised after seeing the others' labels. Revision behavior varied across labelers, and the human
  annotator's revisions frequently introduced framings absent from the ensemble's collective output. We argue disagreement-based
  evaluation is a necessary complement to accuracy metrics for LLM-assisted interpretive coding.

\end{abstract}

\section{Introduction}

Large language models are being used to summarize and categorize public comments, but prevailing evaluation methods cannot assess     
  whether these interpretations faithfully represent what the public submitted. This gap matters because when LLM outputs characterize
  public input, they produce structured representations that shape what decision-makers perceive as salient and what they believe the   
  record contains. Yet standard accuracy metrics can create an appearance of validity while leaving representational validity           
  unexamined.                                                                                                                           
                                                            
  The primary
   mechanism for this input is notice-and-comment rulemaking, where agencies solicit public responses to proposed regulations. The
  submitted comments and the agency's responses form the administrative record, the formal evidentiary basis on which the final rule
  must stand. Research demonstrates that these comments shape policy outcomes: agencies routinely adjust proposed rules in response to
  the tone and substance of submitted comments \cite{yackee2006sweet, kirilenko2014us}. But agencies face a persistent bottleneck: they
  must translate large, messy corpora of public input into tractable representations, often on short timelines and under staffing
  constraints.

  Although LLM hallucinations motivate careful evaluation in high-stakes applications \cite{gallagher2025evaluating,                    
  gennari2024considerations}, public comment analysis presents a distinct challenge. The risk extends beyond misstating facts: LLMs
  shape the representation of the record itself, including what themes exist, which arguments matter and how diverse perspectives are   
  preserved or collapsed. Different models can produce coherent outputs that still disagree in how they organize the same corpus.
  Standard evaluation privileges what is easiest to score, such as stance accuracy on a labeled subset, and treats the rest of the
  interpretive pipeline as secondary \cite{fu2024deciphering}. A common response is expert-led prompt iteration and small-sample
  validation \cite{deiner2024large,HHS2024LLMCommentPilotLessons}. But for interpretive tasks, this often amounts to choosing among
  plausible readings rather than verifying against a stable ground truth \cite{kuo2023understanding,showkat2023right}. 
  
  We build on these works by rethinking how LLM-powered analysis can be made more accountable to the diversity of the public record.    
  Rather than selecting one model based on accuracy alone, we propose an Interpretive Audit Pipeline that treats multi-model
  disagreement as an analytical instrument. 

Categorization schemes are analytical choices with material consequences~\citep{scott1998seeing,saxena2024algorithmic}. Current       
  evaluation checks stance accuracy on a labeled subset without asking whether the model's thematic organization preserves the diversity
   of arguments in the record. 
Our results show that this gap is  
  consequential: models that score similarly on stance produce materially different thematic outputs. We use \textit{representational
  validity} to ask whether a model preserves the diversity of arguments in the record, rather than compressing or fragmenting
  it~\cite{selbst2019fairness,stapleton2022imagining}, and \textit{interpretive disagreement} for cases where models look similar on a scorable subtask but diverge in how they organize themes. We develop this argument empirically using 1,260 public comments submitted to USDA docket FNS-2016-0018 on SNAP retailer stocking    requirements. We compare four LLMs (Gemini-3.1-pro, GPT-5.4, Llama-3.3-70B, Mistral-Medium) across five open-extraction prompt variants and a closed-extraction workflow using an expert nutritionist rubric, and conduct a stratified 40-comment two-stage labeling study with a human annotator. We show that: (1) under fixed model versions and default settings, between-model variation in
  distributional metrics exceeds within-model prompt variation; (2) an expert rubric eliminates deep interpretive disagreement not by
  resolving it but by suppressing it; (3) the disagreement structure across models is itself informative, mapping the interpretive
  complexity of the corpus and identifying where the public record is genuinely multi-argument; and (4) in the labeling study, revision
  behavior varies across labelers and the human annotator's revisions frequently introduce framings absent from the LLM ensemble's
  collective output. 
  We argue the ML community must develop
  disagreement-based evaluation methods that make interpretive divergence measurable and
  contestable~\cite{karusala2024understanding,green2022flaws}.

\section{Related work}
 
\paragraph{LLMs for public comment and policy analysis.}
A growing body of work applies LLMs to public comment corpora, motivated by the administrative
bottleneck of processing large volumes of citizen input on short timelines. \citet{fu2024deciphering}
benchmark ChatGPT against human coders for citizen feedback analysis, finding competitive
performance on stance but noting limitations in thematic consistency. \citet{rong2025benchmarking}
compare LLMs and NLP methods for decoding public sentiment on urban policy, emphasizing speed and
operational feasibility. \citet{lopez2025surfacing} propose an LLM-based method for surfacing citizens' policy perspectives from open-ended responses, using concept extraction and semantic clustering. \citet{deiner2024large} assessed whether LLMs can perform inductive thematic analysis of social media corpora in a single prompt, finding potential for automated social listening but noting that LLMs did not 
consistently replicate the themes identified by human experts, proposing human validation
as a quality check. Tools such as PolicyPulse~\cite{wang2025policypulse} demonstrate that
LLM-powered systems can help researchers surface
unexpected themes, with the explicit goal of capturing diverse societal perspectives that
traditional data sources miss. Government practice is moving in the same direction: the HHS pilot
on public comment analysis~\cite{HHS2024LLMCommentPilotLessons} centers responsible use on
human-in-the-loop review. Our work departs from this literature by treating the \textit{variation}
across models and prompts as an object of analysis.\\
\textbf{Evaluation beyond accuracy.}
Standard evaluation of LLMs on classification tasks relies on agreement with a human-labeled
subset, typically reported as accuracy or F1. This framing has well-documented limitations in
high-stakes interpretive settings. \citet{selbst2019fairness} show that fairness constraints can become meaningless when abstracted from their sociotechnical context; we extend this concern to thematic coding, where accuracy on stance can coexist with systematic misrepresentation of
argument structure. \citet{stapleton2022imagining} and \citet{saxena2024algorithmic} document how
algorithmic categorization in public services embeds contested value choices that accuracy metrics do not surface. \citet{karusala2024understanding} examine contestability on the margins of
algorithmic decision-making, arguing that evaluation methods must make the stakes of disagreement visible rather than averaging over them. \citet{green2022flaws} critiques human-oversight requirements as insufficient when the oversight is not equipped to detect distributional errors.
Our disagreement taxonomy operationalizes these concerns. We ask where models diverge and what that divergence reveals about the corpus.\\
\textbf{Perspectivist annotation.}
A line of work in NLP and HCI argues that disagreement among annotators is signal about the data rather than noise to be aggregated away.  \citet{plank2022problem} reviews the case for distribution-aware   modeling: where humans systematically disagree, treating one labeler's reading as ground truth and scoring the rest as error mismeasures the underlying phenomenon. \citet{kapania2023hunt} extend this to annotation labour, showing that the ``ground truth'' framing relies on representationalist assumptions about whether data can be neutrally                captured, and that annotation is interpretive work whose plurality                 should be preserved. \citet{karusala2024understanding} extend the                  concern further to the design of algorithmic decision-making in public             services: contestability, the ability of those affected by algorithmic             decisions to follow up, appeal, or challenge them, requires that                   classification be visible and challengeable rather than appearing as a             single authoritative output. Earlier work on crowd annotation                      \citep{aroyo2015truth} and recent perspectivist evaluation efforts                 \citep{basile2021need, davani2022dealing, prabhakaran2021releasing}                share this commitment. We extend this lineage to LLM-based labeling:               model identity and prompt phrasing introduce labeler-like variation,               and preserving that variation as distinct labels, each retained                    alongside the model that produced it, is what makes the LLM-assisted               categorization legible to a downstream reviewer and, in principle,                 contestable. The human annotator in our 40-comment study is positioned             as one interpretive reader producing labels in parallel with the LLMs rather than as a downstream validator of model output.\\ 
\textbf{The multi-reader model.}
The closest intellectual precedent for our study is not in ML but in qualitative methods. In grounded theory \citet{strauss1990basics}, multiple coders independently develop codes before negotiating consensus, precisely because disagreement surfaces interpretive possibilities that any single pass would foreclose. We draw on \citet{deane2020building}, 
whose work suggests that human interpretation is fundamentally recursive, to motivate the research design, not to claim that LLMs possess interpretive perspectives analogous to human readers. The empirical 
question is whether cross-model variation in thematic output is structured and informative.
\citet{scott1998seeing} provided the political theory grounding: state administrative systems require simplification of complex phenomena into standard categories, and these simplifications are not neutral, they determine what becomes visible in the administrative record. Taken together, these traditions motivate our approach: rather than selecting a single model based on accuracy alone, we compare multiple models' outputs, characterize the structure of their divergence, and assess 
whether that structure can support human interpretive work.\\
\textbf{Multi-LLM evaluation and aggregation.}
  A growing body of work uses multiple LLMs jointly for evaluation, generation, or qualitative coding. ChatEval \citep{chan2023chateval}, PoLL \citep{verga2024replacing}, and LLM-as-jury approaches aggregate multiple LLM judgments to improve a  single output. Multi-agent debate frameworks \citep{du2023improving, liang2024encouraging} have models exchange critiques to converge on a shared answer. \citet{chen2023multiagent} and \citet{borchers2025temperature} extend this convergence-seeking approach to       numerical and qualitative-coding settings respectively. Pipelines such as Thematic-LM \citep{thematiclm2024} merge multi-agent codings into a single shared codebook. \citet{tajik2026disagreement} reduce reasoning-trace disagreement to a scalar metric for downstream 
  review. Each of these resolves cross-LLM disagreement (by arbitration, aggregation, debate-to-consensus, or scalar reduction) rather
  than preserving it. \citet{kambhampati2025stop} caution against anthropomorphizing such exchanges as deliberation. Our approach diverges in goal: rather than resolving disagreement, we preserve per-model labels as distinct readings, each retained alongside the model that produced it, and treat the resulting disagreement structure as the artifact a human reviewer engages with.
\vspace{-8pt}
\section{Methods}
We designed an experiment to test whether, under fixed model versions and default settings, differences across LLMs matter for thematic coding of public comments, and whether the resulting disagreement structure can support human interpretation. %

\subsection{Dataset}

We analyzed 1,260 public comments submitted to USDA Food and Nutrition Service Docket FNS-2016-0018 concerning SNAP retailer stocking requirements, a rule affecting about 260,000 retailers and 47 million beneficiaries \cite{Oliveira2018SNAP} at the time of the rule change. Data collection and preprocessing details are reported in Appendix~\ref{app:preprocessing}. 

\subsection{LLM Experimental design}
\label{sec:llm-prompts}

We selected four models: Gemini-3.1-pro, GPT-5.4, Llama-3.3-70B-Instruct, and Mistral-Medium-2505. Prompts are identically worded across all models and run at each model's default settings; full prompt texts are given in Appendix~\ref{app:prompts}.

\textbf{Prompt 1 (P1):} Prompt 1 specifies an output schema for each comment: stance (Support / Oppose), main argument (1--2 words), one-sentence explanation, submitter type, and confidence rating.

To assess sensitivity to prompt wording, we developed five variants holding the task and output schema constant.  \textbf{Verbosity} uses a minimal instruction with no elaboration. \textbf{Format} restructures the same task using explicit section headers and bullet-point formatting. \textbf{Lexical} substitutes near-synonyms throughout (e.g., \textit{categorize} for \textit{classify}, \textit{central issue} for \textit{primary concern}). \textbf{Role} defines a persona, a senior federal policy analyst, before an otherwise standard instruction. \textbf{Binary} provides per-field rationale with examples to test whether more prescriptive scaffolding shifts classification behavior. Each variant was run once per model (20 runs total: 4 models $\times$ 5 variants).

\textbf{Prompt 2 (P2):} Asks the models to map comments to one of 17 categories in a predefined human-expert rubric \citep{haynes2018arguments}, returning stance, main argument, submitter type, and confidence. To estimate variance we ran this prompt ten times per model at default settings (40 runs total: 4 models $\times$ 10 replications). %

\subsection{Human expert baseline}

In previously published research \cite{haynes2018arguments}, three researchers and domain experts independently developed a 
rubric with stance categories (Support / Oppose) and thematic arguments (economic impact, food access, nutrition/health, program integrity, etc.) for the same dataset using 303 comments selected through random sampling. This is the rubric that was given to the LLMs for Prompt 2, and the one we treated as the gold standard. %
Because the original publication did not report which 303 comments were selected,
we could not match their stance labels directly. To establish a validation set, we independently
sampled and classified 150 comments, which we use for all stance and theme accuracy evaluations.

\subsection{Disagreement taxonomy}
\label{sec:taxonomy}
 
To characterize cross-model divergence beyond aggregate metrics, for each comment, we first identify which cluster each model assigned most often across its runs. This gives one vote per model (four votes total per comment).
We classify each comment into one of four disagreement types: \textit{Consensus} (all four model votes assign the same cluster),    
  \textit{Binary theme split} (models divide into exactly two distinct clusters), \textit{Deep theme disagree} (three or four distinct  
  clusters appear across the four model votes; the comment resists single-frame classification), or \textit{Stance disagree} (models
  disagree on Support vs.\ Oppose regardless of thematic assignment).

This taxonomy treats disagreement as informative about the corpus rather than as measurement error: a comment classified as \textit{Deep theme disagree} is one where the text is doing multiple argumentative jobs simultaneously, and no single cluster label is adequate.

\subsection{Human label augmentation: comparing LLMs and human recoding}
\label{sec:revision-protocol}                  To assess whether multi-model disagreement structure can support human interpretive work, we conducted a two-stage comparison of LLM and human labeling on a stratified 40-comment subsample.                       %
\textbf{Stratified sample.}We constructed the 40-comment subsample to span the corpus's interpretive structure. The sample was balanced across the four
disagreement-taxonomy categories defined in Section~\ref{sec:taxonomy} (Consensus, Binary theme split, Deep theme disagree, Stance    
disagree), each derived from the rubric-guided modal classification of every comment in the corpus. Stratification was crossed with
two source-type strata (text in the comment field vs.\ text in an attached document) to ensure coverage of both short opinion comments
and longer document-style submissions. Selection used a fixed random seed within strata; form-letter campaigns were de-duplicated
prior to sampling and comments with unreadable text were excluded. Full selection logic, exclusion sets, de-duplication procedure, and
seed value are detailed in Appendix~\ref{app:stratified-40}.

\textbf{First stage --- independent labeling.}                                        
Each of the four LLMs classified each of the 40 comments using the Verbosity variant prompt (Section~\ref{sec:llm-prompts}) at temperature 1.0, producing one label and one evidence quote per comment. A human annotator labeled the same 40 comments in parallel, without observing any model output, also producing a label and a short evidence quote per comment.

\textbf{Second stage --- revision under additional context.}                          
Each LLM was re-prompted with the comment plus the four first-stage LLM labels (its own and the three other models'), shuffled, with
its own reading marked, and produced a revised label and evidence quote. The annotator was shown the deduplicated union of all eight  
LLM labels (four first-stage + four second-stage) per comment, showing which models surfaced each concept along with one
representative evidence quote per concept, and decided whether to keep the first-stage label, change to one shown, or generate a      
different label.                                          
 \textbf{Outcome categories.} For each labeler we classify the second-stage label relative to the first-stage labels: \textit{hold} 
  (clusters with the labeler's own first-stage label), \textit{absorb} (clusters with another labeler's first-stage label but not the   
  labeler's own), or \textit{novel} (clusters with neither). We cluster LLM and annotator labels jointly. Details are in                
  Appendix~\ref{app:clustering}.

Together, the disagreement taxonomy and the human label augmentation protocol constitute the Interpretive Audit Pipeline. Rather than 
selecting a single model based on accuracy, this pipeline directs evaluation of LLM-assisted interpretive coding to compare outputs across models, characterize the structure of their divergence, and 
direct human interpretive labor toward comments where that divergence persists.

\subsection{Evaluation metrics}
We evaluate LLM outputs using traditional classification accuracy and structural metrics: Shannon entropy (evenness of theme distribution), Gini coefficient (concentration across themes), and Jaccard similarity (overlap between theme vocabularies). Definitions and details are included in Appendix \ref{app:metrics}. Together these metrics serve as descriptive diagnosis to hint at whether a model captures diverse public input or collapses variation into dominant themes, and whether that variation is driven by model type or prompt surface form.

\section{Results}
We present findings in three parts: (1) classification performance under P1, (2) the effect of introducing an expert rubric under P2 and (3) the parallel labeling study on a stratified 40-comment subsample. %
\subsection{Open-ended classification performance - P1}
\label{sec:p1results}
Theme generation varied substantially across models (Table~\ref{tab:prompt1_results}). Gemini and GPT-5 generated most unique themes $232 \pm 27$ and $237 \pm 39$ respectively, roughly $43\%$ more than Llama and Mistral $162 \pm 26$ and $137 \pm 9$. This split is similar in the distributional metrics. Entropy was higher for Gemini and GPT-5 ($6.00 \pm 0.31$ and $5.99 \pm 0.49$ bits), indicating more even dispersal of comments across themes, while Llama and Mistral ($5.30 \pm 0.32$ and $5.16 \pm 0.17$ bits) showed greater concentration in fewer dominant categories. Gini coefficients followed the same ordering, though the separation between models is more modest than with entropy.

Despite these differences in thematic granularity, stance accuracy was high and stable across all models ($92.67$--$95.60\%$) indicating that support/oppose classification is largely robust to both model choice and prompt surface form. 
 
\begin{figure}[t]
    \centering
    \includegraphics[width=.9\linewidth]{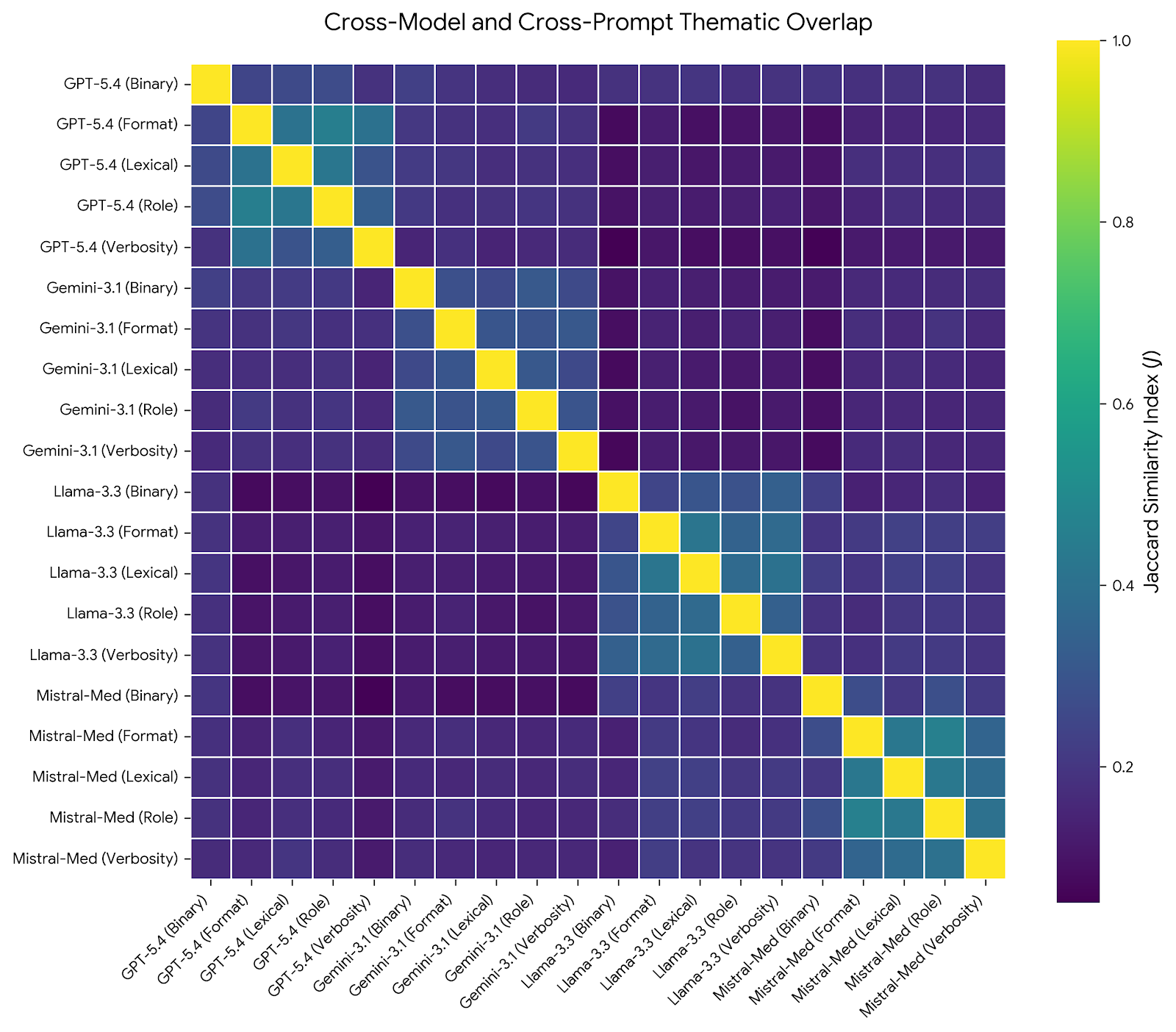}
    \caption{Jaccard Similarity Matrix for cross-model thematic overlap. The heatmap shows higher model similarity across different prompt variations than across models.}
    \label{fig:jaccard}
\end{figure}

To disentangle the relative contribution of model type versus prompt formulation, we computed pairwise Jaccard similarity across all 20 model, prompt combinations in P1 (Figure~\ref{fig:jaccard} within-model blocks exhibit higher similarity, lighter teal values, while cross-model regions remain dark blue.) Same-model cross-prompt similarity averages $0.13$ (range $0.01-0.25$) while cross-model similarity averages $0.03$, with 97\%  of pairs below $0.08$. Because Jaccard operates on raw label strings, this gap may partly reflect vocabulary differences rather than organizational divergence, however, the 
embedding-based clustering in Section~\ref{sec:results-parallel}, which merges paraphrases into shared clusters, produces a consistent picture, models still land in distinct clusters even after surface variation is controlled for. Taken together, these results suggest that model identity induces greater divergence in thematic vocabulary and organization than prompt variation, though the relative magnitude of the two components cannot be precisely separated with Jaccard alone.
 
The disagreement taxonomy (Section~\ref{sec:taxonomy}) reveals the comment-level structure behind these aggregate patterns. Under P1, only $1.4\%$ of comments reached full consensus across all four models. The remaining exhibited some form of interpretive divergence: $13.9\%$ binary theme splits, $78.3\%$ deep theme disagreements (three or four distinct framings), and $6.3\%$ stance disagreements.

\begin{table*}[t]
\centering
\small
\begin{tabular}{ll ccccc c}
\toprule
\textbf{Metric} & \textbf{Model} & \textbf{Verbosity} & \textbf{Role} & \textbf{Format} & \textbf{Lexical} & \textbf{Binary} & \textbf{Mean $\pm$ SD} \\
\midrule
\multirow{4}{*}{\textit{Themes}}
 & Gemini-3.1-pro     & 237 & 211 & 204 & 233 & 274 & $232 \pm 27$ \\
 & GPT-5.4            & 288 & 248 & 253 & 200 & 196 & $237 \pm 39$ \\
 & Llama-3.3-70B      & 155 & 190 & 178 & 165 & 122 & $162 \pm 26$ \\
 & Mistral-Medium     & 137 & 139 & 145 & 141 & 122 & $137 \pm 9$ \\
\midrule
\multirow{4}{*}{\textit{Accuracy (\%)}}
 & Gemini-3.1-pro     & 94.00 & 94.00 & 92.67 & 94.00 & 92.67 & $93.47 \pm 0.73$ \\
 & GPT-5.4            & 93.33 & 92.00 & 92.67 & 92.67 & 92.67 & $92.67 \pm 0.47$ \\
 & Llama-3.3-70B      & 92.67 & 94.67 & 95.33 & 94.67 & 94.00 & $94.27 \pm 1.01$ \\
 & Mistral-Medium     & 95.33 & 96.00 & 96.00 & 96.67 & 94.00 & $95.60 \pm 1.01$ \\
\midrule
\multirow{4}{*}{\textit{Gini}}
 & Gemini-3.1-pro     & 0.697 & 0.716 & 0.726 & 0.731 & 0.679 & $0.710 \pm 0.022$ \\
 & GPT-5.4            & 0.685 & 0.697 & 0.708 & 0.755 & 0.762 & $0.721 \pm 0.035$ \\
 & Llama-3.3-70B      & 0.757 & 0.729 & 0.754 & 0.724 & 0.782 & $0.749 \pm 0.023$ \\
 & Mistral-Medium     & 0.765 & 0.769 & 0.785 & 0.753 & 0.753 & $0.765 \pm 0.013$ \\
\midrule
\multirow{4}{*}{\textit{Entropy (bits)}}
 & Gemini-3.1-pro     & 6.13 & 5.89 & 5.76 & 5.75 & 6.49 & $6.00 \pm 0.31$ \\
 & GPT-5.4            & 6.47 & 6.34 & 6.20 & 5.51 & 5.43 & $5.99 \pm 0.49$ \\
 & Llama-3.3-70B      & 5.16 & 5.63 & 5.39 & 5.51 & 4.81 & $5.30 \pm 0.32$ \\
 & Mistral-Medium     & 5.23 & 5.16 & 4.88 & 5.32 & 5.21 & $5.16 \pm 0.17$ \\
\bottomrule
\end{tabular}
\caption{P1 results across five prompt variants. Each cell is a single run
(one model $\times$ one prompt variant). The Mean $\pm$ SD column summarizes within-model
variation across prompt surface forms. Accuracy measured against 150 human-labeled comments.
Entropy and Gini are computed over all 1,260 comments.}
\label{tab:prompt1_results}
\end{table*}

\subsection{Impact of structured guidance - P2}

The introduction of a predefined classification rubric from \cite{haynes2018arguments} (P2) narrowed the thematic output of the systems. Unique theme counts decreased from a range of $137\text{--}237$ under P1 prompts to $15\text{--}16$ labels across all four models. This shift is reflected in the Shannon entropy, which decreased from a range of $5.16\text{--}6.00$ bits in the open condition to $2.24\text{--}2.45$ bits in P2. Despite the reduction in label variety, Gini coefficients remained similar to the P1, indicating that the underlying concentration of theme assignments was not affected by the change in prompt structure.

Table \ref{tab:divergence_comparison} summarizes the shifts in disagreement taxonomy. While consensus increased and deep theme disagreement decreased comments continued to exhibit theme-level divergence across both conditions. %

\begin{table}[h]
\centering
\begin{tabular}{@{}lrrrr@{}}
\toprule
\textbf{Class} & \textbf{P1} & \textbf{\%} & \textbf{P2} & \textbf{\%} \\ \midrule
Consensus           & 18  & 1.4\%  & 377 & 29.9\% \\
Binary theme split  & 175 & 13.9\% & 617 & 49.0\% \\
Deep theme disagree & 987 & 78.3\% & 156 & 12.4\% \\
Stance disagree     & 80  & 6.3\%  & 110 & 8.7\%  \\ \bottomrule
\end{tabular}
\caption{Comparison of Disagreement Taxonomy Across P1 and P2 Prompt Conditions}
\label{tab:divergence_comparison}
\end{table}

While stance accuracy remained between $91\%$ and $93\%$, thematic accuracy varied between $41\%$ and $70\%$ across the four models. The models that generated the highest number of themes in P1 prompting variations did not consistently demonstrate higher accuracy under the structured rubric.

\begin{table}[t]
\centering
\small
\begin{tabular}{lrrrr}
\toprule
 & \textbf{Gemini} & \textbf{GPT-5} & \textbf{Llama} & \textbf{Mistral} \\
\midrule
Themes               & $16 \pm 1$ & $15 \pm 1$ & $15 \pm 1$ & $16 \pm 1$ \\
Stance Accuracy      & $0.93 \pm 0.00$ & $0.93 \pm 0.00$ & $0.91 \pm 0.01$ & $0.92 \pm 0.00$ \\
Theme Accuracy       & $0.70 \pm 0.02$ & $0.63 \pm 0.01$ & $0.55 \pm 0.03$ & $0.41 \pm 0.02$ \\
Gini                 & $0.76 \pm 0.01$ & $0.71 \pm 0.01$ & $0.74 \pm 0.02$ & $0.76 \pm 0.01$ \\
Entropy              & $2.24 \pm 0.05$ & $2.45 \pm 0.02$ & $2.33 \pm 0.03$ & $2.32 \pm 0.05$ \\
\bottomrule
\end{tabular}
\caption{P2 metrics, mean $\pm$ SD 
across 10 runs.}
\label{tab:p2_results}
\end{table}

\subsection{Parallel labeling on a stratified sample}
\label{sec:results-parallel}

We applied the parallel labeling procedure described in Section~\ref{sec:revision-protocol} to the stratified 40-comment sample. Four LLMs and one annotator produced labels independently in round 1 and revised after exposure to the others' labels in round 2.

Table~\ref{tab:hold-absorb-novel} reports the distribution of revision outcomes. GPT and Gemini revised in 22.5\% of comments, with no revisions producing framings absent from the round-1 set. Their aggregate distributions are identical, but they disagree on outcome for 12 of 40 comments, indicating distinct revision behavior despite the matching marginals. Mistral and Llama both revised in 42.5\%, but with different distributions: Mistral's revisions clustered most often with another LLM's round-1 label (paraphrastic absorption), while Llama's revisions more often produced new framings. The annotator revised in 52.5\% of comments; 12 of the 21 revisions clustered with framings already in the LLM round-1 set, and 9 produced new framings. These counts describe revision behavior under exposure; they do not measure label quality. The four cases below illustrate what these revisions look like in practice.

\begin{table}
\centering
\begin{tabular}{lrrrr}
\toprule
Labeler & Hold & Absorb & Novel & Change rate \\
\hline
GPT-5.4 & 31 & 9 & 0 & 22.5\% \\
Gemini-3.1-pro & 31 & 9 & 0 & 22.5\% \\
Mistral-Medium & 23 & 11 & 6 & 42.5\% \\
Llama-3.3-70B & 23 & 5 & 12 & 42.5\% \\
Annotator & 19 & 12 & 9 & 52.5\% \\
\hline
\end{tabular}
\caption{Round-2 revision outcomes for the four LLMs and the annotator. \textit{Hold}: round-2 label clusters with the labeler's own round-1 label. \textit{Absorb}: round-2 label clusters with another labeler's round-1 label. \textit{Novel}: round-2 label clusters with neither.}
\label{tab:hold-absorb-novel}
\end{table}

\textbf{FNS-2016-0018-0944.} The comment is from a convenience store worker writing in opposition: \textit{``We respond to their   
  changing demands by stocking more and more fresh and healthy foods than we have in the past. But SNAP beneficiaries are not our only  
  customers therefore we also offer a selection of hot prepared foods, non food items and other products that are not apart of SNAP   
  benefits. \dots\ FNS should not require a certain number of units that need to be displayed for each product at all times. This will  
  result in waste for our perishable food items that will spoil.''}

The four LLMs land on different framings in round 1: \textit{small store burden} (GPT), \textit{Inventory waste} (Gemini), \textit{Small Business Impact} (Llama), \textit{small business impact} (Mistral). In round 2, GPT and Gemini hold; Mistral revises to \textit{inventory requirements}; Llama revises to \textit{Practicality Concerns}. The annotator's round-1 label is \textit{Rules will restrict convenience stores' ability to operate}, revised in round 2 to \textit{Unworkable requirements for convenience stores}. The annotator notes that ``the unworkable requirements label identified by the LLM helped guide the revision.'' Across the ten readings, six distinct framings remain after round 2.

 \textbf{FNS-2016-0018-0103.} The comment is from a dietetic intern writing in support: \textit{``I highly support this proposal as 
  there are millions of families across the United States who are unable to better their health and prevent disease through nutrition   
  due to financial limitations and lack of access to healthy foods.''}

GPT and Gemini both label this as \textit{healthy food access} in round 1 and hold or paraphrase in round 2. Llama moves from \textit{Health Benefits} to \textit{Nutrition Access}. Mistral moves from \textit{healthier food access} to \textit{food insecurity reduction}. The annotator labels it \textit{Proposal improves health of less-affluent people} in round 1 and revises to \textit{Proposal improves population health outcomes}, noting that an LLM evidence quote ``showed me that the previous label had been too constricted to less-affluent people.'' Five distinct framings remain after round 2; none of the LLMs reach the annotator's framing in either round.

\textbf{FNS-2016-0018-0085.} The comment is an attached letter from a dietetic intern in rural Wisconsin: \textit{``Food insecurity
   is a growing problem in the U.S. \dots\ many people in the county live in very rural areas with limited access to stores and         
  transportation. \dots\ Using their SNAP benefits these families are able to shop at local mom and pop type grocery stores and       
  convenience stores that are a SNAP retailer.''}

The LLMs spread across access framings in round 1: \textit{access and nutrition} (GPT), \textit{Improves participant nutrition} (Gemini), \textit{Rural Access} (Llama), \textit{Need for access} (Mistral). In round 2 they converge toward access-related framings (\textit{Need for access}, \textit{Rural Access}, \textit{Access and nutrition}). The annotator labels the comment \textit{Rule changes accelerate food insecurity} in both rounds, noting that ``food insecurity is slightly more encompassing than the category of food access that the LLMs are using.'' The annotator's framing does not appear in the LLM set at either round.

\textbf{FNS-2016-0018-0021.} The comment is short and supports the rule: \textit{``I strongly support this rule to ensure that a
  wider variety of fresh, whole and minimally processed foods, especially fruits and vegetables be made regularly available in all food 
  retail establishments that accept SNAP vouchers. In areas known as `food deserts' this is monumentally important.''}

In round 1, all four LLMs label this in the healthy-access region (\textit{healthy access}, \textit{Food Access}, \textit{Healthy food access}, \textit{Healthy food access}). In round 2, GPT and Gemini hold; Mistral revises to \textit{Food deserts}; Llama revises to \textit{Food Equity}. The annotator labels it \textit{Increased Access to Healthy Food} in round 1 and revises to \textit{Rules increase healthy food accessibility} in round 2, holding the same cluster. The number of distinct framings increases from two in round 1 to four in round 2.

Across the four cases, several patterns appear. The same comment often supports more than one reasonable framing, and labels that differ in wording are not always wrong relative to each other. Exposure to other labels does different things in different cases: it can sharpen a labeler's wording (0944), broaden it (0103), or produce additional distinct framings rather than convergence (0021). In some cases the annotator's framing sits inside the LLM set (0944 round 2, 0021); in others it sits outside (0085, 0103). We return to the methodological implications in Section~\ref{sec:discussion}.

\section{Discussion}
  \label{sec:discussion}                               This paper asks how LLM outputs should be evaluated when the task is interpretive coding rather than classification against a stable answer. Standard evaluation reports accuracy against a small reference labeling. For thematic coding of public comments, the same comment often supports several reasonable readings, and a single accuracy number cannot surface this. Even with an expert rubric, more than half of the comments retained multiple reasonable framings. %
  Label variation in interpretive tasks is information about the data, not noise to be averaged out \citep{plank2022problem,aroyo2015truth}.

  Stance accuracy stayed within 91--96\% across models and prompts, yet the same models produced different thematic outputs. %
  Stance accuracy measures whether the model picked Support or Oppose, not how the model organized the comment thematically. The thematic organization shapes how a policymaker reads the record. 
  Most multi-LLM evaluation methods aggregate across models to produce a single output \citep{chan2023chateval,verga2024replacing,du2023improving}; we argue the opposite. An evaluation pipeline for interpretive coding should retain each labeler's reading alongside the labeler that produced it, so that alternatives remain visible and 
contestable~\citep{karusala2024understanding,prabhakaran2021releasing,davani2022dealing}.

   In our study, the human annotator reads alongside the LLMs rather than validating their output, \citep{kapania2023hunt}; the aggregation, if needed, happens at the reviewer's end.
This study deliberately avoids defining a normative target for thematic organization. The absence is not an oversight but a consequence of the argument: for interpretive tasks where the same text supports multiple reasonable framings, no single organization is correct. The evaluation contribution is not a scoring function but a diagnostic structure,  disagreement type, distributional metrics, and revision. That makes the interpretive consequences of model choice visible to a human reviewer. The reviewer, not the evaluation pipeline, decides which organization best serves the administrative purpose at hand.

  The pipeline positions LLMs as instruments that augment human interpretation under resource constraints. Agencies process large comment corpora on short timelines with limited staff. %
The claim is not that four models is the right number; it is that one model, however accurate on a labeled subset, creates a mirage of thematic completeness. Evaluation for LLM-assisted public-comment analysis should report distributional metrics alongside stance accuracy, surface disagreement taxonomies across models and prompts, and identify comments where interpretive variation persists, directing human reading where the record demands it.

\section{Limitations}
This study has several limitations. The analysis covers a single SNAP regulation docket whose technical and repetitive nature may not generalize to more polarized or differently structured policy domains. Proprietary models are updated regularly; our results represent a snapshot under fixed versions and default settings, and behavior may shift across updates. The 
40-comment study relies on one human annotator, and the 150-comment validation set represents one reading of the corpus; multiple coders with varied backgrounds would strengthen claims about the boundary between 
LLM-reachable and human-only framings. The Jaccard analysis operates on raw label strings and may partly reflect vocabulary differences rather than organizational divergence; the embedding-based clustering in Appendix~\ref{app:clustering} provides partial but not complete control for this. The hold/absorb/novel classification depends on a clustering distance threshold with no principled selection method beyond inspecting 
coherence. Finally, we tested two sources of within-model variation and found neither approached 
between-model divergence. It is possible that cumulative within-model variation could 
narrow the gap under non-default settings.

\begin{ack}
This paper presents work whose goal is to advance the field of machine learning by improving evaluation practices for interpretive LLM tasks in high-stakes policy settings. There are potential societal consequences of our work, including influences on public comment analysis and administrative decision-making, none of which we feel must be specifically highlighted here.
\end{ack}

\newpage
\bibliography{example_paper}
\bibliographystyle{plainnat}

\appendix
\section {Data Pre-processing Details}
\label{app:preprocessing}

The full dataset of 1,260 public comments on docket FNS-2016-0018 was collected using the Regulations.gov API (v4) and custom web scraping tools. The Regulations.gov API v4 does not provide attachment URLs or file download endpoints, and the website uses JavaScript rendering which prevents simple HTTP-based scraping of attachment links. We discovered (via Claude Code) that attachments were publicly accessible via predictable URL structures once the correct pattern was identified. 

\section{Prompts}
\label{app:prompts}
\subsection{Prompt 1}

Below the five prompt variations where the LLM were instructed to classify the public comments
\subsubsection{Format}
SYSTEM\_PROMPT = You will classify public comments on a proposed regulation. First, read this proposed regulation. Then, read the public comments.

PROPOSED RULE CONTEXT: \{\{RULE\_TEXT\}\}

CLASSIFICATION TASK:

For each comment, provide these 5 classifications:

\textbf{Support/Oppose}
\begin{itemize}
  \item Does the commenter support or oppose the rule?
  \item Read the ENTIRE comment before deciding
  \item Note: ``do not support'' = Oppose
  \item Values: Support $|$ Oppose
\end{itemize}

\textbf{Main Argument}
\begin{itemize}
  \item The commenter's primary concern in 1--2 words
\end{itemize}

\textbf{Argument Expanded}
\begin{itemize}
  \item The argument explained in one sentence
\end{itemize}

\textbf{Submitter Type}
\begin{itemize}
  \item Values: Business $|$ Nonprofit $|$ Individual $|$ Government $|$ Unknown
  \item Check the organization field first, then the comment text
\end{itemize}

\textbf{Confidence}
\begin{itemize}
  \item How much textual evidence supports your classifications?
  \item High: position is clear
  \item Medium: Some inference is needed
  \item Low: ambiguous or insufficient text
  \item Note: rate evidence quality, not your internal certainty
\end{itemize}

OUTPUT FORMAT: Respond ONLY with a valid JSON array. No preamble, no explanation, no markdown. Each element must have exactly these keys:

\begin{verbatim}
[
  {
    "comment_index": 0,
    "support_oppose": "value",
    "main_argument": "value",
    "argument_explanation": "value",
    "submitter_type": "value",
    "confidence": "value"
  }
]
\end{verbatim}

\subsubsection{Lexical}
SYSTEM\_PROMPT = You will categorize public comments on a pending rule.

PROPOSED RULE CONTEXT: \{\{RULE\_TEXT\}\}

CLASSIFICATION TASK:

For each comment, provide these 5 categorizations:
\begin{enumerate}
  \item \textbf{Support/Oppose:} Does the commenter support or oppose the rule? Read the ENTIRE comment first. ``do not support'' = Oppose. Output: Support, Oppose
  \item \textbf{Main Argument:} What's their central issue? (1--2 words)
  \item \textbf{Argument Expanded:} Explain the argument in one sentence
  \item \textbf{Submitter Type:} Business, Nonprofit, Individual, Government, or Unknown. Check organization field first, then comment text.
  \item \textbf{Confidence:} How much textual support backs these categorizations? High = position is clear, Medium = some inference needed, Low = ambiguous or insufficient text. Confidence = evidence in text, not how certain you feel.
\end{enumerate}

You will receive multiple comments at once.

OUTPUT FORMAT: Respond ONLY with a valid JSON array. No preamble, no explanation, no markdown. Each element must have exactly these keys:

\begin{verbatim}
[
  {
    "comment_index": 0,
    "support_oppose": "value",
    "main_argument": "value",
    "argument_explanation": "value",
    "submitter_type": "value",
    "confidence": "value"
  }
]
\end{verbatim}

\subsubsection{Binary}
You are tasked with classifying public comments on a proposed federal regulation.

PROPOSED RULE CONTEXT: \{\{RULE\_TEXT\}\}

CLASSIFICATION TASK:

Read each comment in full and assign the following five fields. Do not use any predefined category list --- derive everything directly from the comment text.

\begin{enumerate}
  \item \textbf{support\_oppose} \\
  Identify the commenter's overall stance toward the proposed rule. \\
  Valid values: ``support'', ``oppose''
  \begin{itemize}
    \item ``support'': commenter approves of or endorses the rule
    \item ``oppose'': commenter disapproves of or objects to the rule
    \item NOTE: ``do not support'' = oppose
  \end{itemize}

  \item \textbf{main\_argument} \\
  Summarize the commenter's primary argument or concern in 1--2 words. \\
  Examples: ``Cost'', ``Nutrition'', ``Implementation'', ``Privacy'', ``Equity'' \\
  Derive this freely from the comment --- do not match to any fixed list.

  \item \textbf{argument\_explanation} \\
  Write 1--2 sentences explaining the commenter's main point in your own words. Be specific to what this commenter actually said.

  \item \textbf{submitter\_type} \\
  Classify who submitted the comment. Use the organization field first; if empty, infer from the comment text. \\
  Valid values: ``Retail/Business'', ``Nonprofit'', ``Private Citizen'', ``Government'', ``Medical Field'', ``Unknown''

  \item \textbf{confidence} \\
  How clearly the comment expresses its stance and argument.
  \begin{itemize}
    \item High = position is clear
    \item Medium = some inference needed
    \item Low = ambiguous or insufficient text
    \item NOTE: Confidence reflects evidence in the text, not internal certainty
  \end{itemize}
\end{enumerate}

OUTPUT FORMAT: Respond ONLY with a valid JSON array. No preamble, no explanation, no markdown. Each element must have exactly these keys:

\begin{verbatim}
[
  {
    "comment_index": 0,
    "support_oppose": "value",
    "main_argument": "value",
    "argument_explanation": "value",
    "submitter_type": "value",
    "confidence": "value"
  }
]
\end{verbatim}

\subsubsection{Role}
SYSTEM\_PROMPT = You are a senior policy analyst at a federal regulatory agency with extensive experience processing public comments. Your role is to systematically analyze and classify public comments submitted in response to a proposed regulation, maintaining objectivity and grounding all judgments strictly in the text of each comment.

PROPOSED RULE CONTEXT: \{\{RULE\_TEXT\}\}

CLASSIFICATION TASK: For each comment, provide these 5 classifications:
\begin{enumerate}
  \item \textbf{Support/Oppose:} Does the commenter support or oppose the rule? Read the ENTIRE comment first. ``do not support'' = Oppose. Output: Support, Oppose
  \item \textbf{Main Argument:} What's their primary concern? (1--2 words)
  \item \textbf{Argument Expanded:} Explain the argument in one sentence
  \item \textbf{Submitter Type:} Business, Nonprofit, Individual, Government, or Unknown. Check organization field first, then comment text.
  \item \textbf{Confidence:} How much textual evidence supports these classifications? High = position is clear, Medium = some inference needed, Low = ambiguous or insufficient text. Confidence = evidence in text, not how certain you feel.
\end{enumerate}

You will receive multiple comments at once.

OUTPUT FORMAT: Respond ONLY with a valid JSON array. No preamble, no explanation, no markdown. Each element must have exactly these keys:

\begin{verbatim}
[
  {
    "comment_index": 0,
    "support_oppose": "value",
    "main_argument": "value",
    "argument_explanation": "value",
    "submitter_type": "value",
    "confidence": "value"
  }
]
\end{verbatim}

\subsubsection{Verbosity}
SYSTEM\_PROMPT = Classify public comments on a proposed regulation.

PROPOSED RULE CONTEXT: \{\{RULE\_TEXT\}\}

CLASSIFICATION TASK: For each comment output:
\begin{enumerate}
  \item \textbf{Support/Oppose:} Support or Oppose
  \item \textbf{Main Argument:} 1--2 words
  \item \textbf{Argument Expanded:} one sentence
  \item \textbf{Submitter Type:} Business, Nonprofit, Individual, Government, or Unknown
  \item \textbf{Confidence:} High, Medium, or Low
\end{enumerate}

You will receive multiple comments at once.

OUTPUT FORMAT: Respond ONLY with a valid JSON array. No preamble, no explanation, no markdown. Each element must have exactly these keys:

\begin{verbatim}
[
  {
    "comment_index": 0,
    "support_oppose": "value",
    "main_argument": "value",
    "argument_explanation": "value",
    "submitter_type": "value",
    "confidence": "value"
  }
]
\end{verbatim}

\subsection{Prompt 2}
You are tasked with classifying public comments on a proposed regulation, specifically FNS-2016-0018.
PROPOSED RULE:\{\{RULE\_TEXT\}\}
\begin{verbatim} 
CODING RUBRIC - 17 CATEGORIES:
1. Support/Oppose: support
   Main Argument: Food access
   Definition: Mentions availability, accessibility (proximity/travel), affordability (price), accommodation (hours/payment types), and acceptability (attitudes) regarding healthy food.

2. Support/Oppose: support
   Main Argument: Access to Healthy Food
   Definition: General belief that the SNAP retailer rule will increase access to healthy food.

3. Support/Oppose: support
   Main Argument: relashionship between nutrition and health
   Definition: Mentions benefits of a healthy diet, consequences of unhealthy eating (chronic disease), and the U.S. obesity epidemic.

4. Support/Oppose: support
   Main Argument: Nutrition
   Definition: Improved nutrition attributable to the SNAP retailer rule.

5. Support/Oppose: support
   Main Argument: Health
   Definition: Improved health outcomes attributable to the SNAP retailer rule.

6. Support/Oppose: support
   Main Argument: Time
   Definition: Comments regarding the time it takes to shop for, prepare, and cook healthy foods.

7. Support/Oppose: support
   Main Argument: Access Associated with Intake
   Definition: Discussion on whether healthy food access is or is not associated with healthy food intake.

8. Support/Oppose: oppose
   Main Argument: SNAP Retailer Rule Definitions
   Definition: Stores' concerns about specific SNAP retailer rule definitions, including retail food stores, multiple food ingredients, and staple foods.

9. Support/Oppose: oppose
   Main Argument: Reduced Food Access
   Definition: Concerns that stores will withdraw from SNAP or close, resulting in reduced food access.

10. Support/Oppose: oppose
   Main Argument: Cost-Benefit Argument
   Definition: Claims the rule hurts businesses by increasing store costs, decreasing profits, and potentially reducing workforces.

11. Support/Oppose: oppose
   Main Argument: Doubting Effectiveness of the rule
   Definition: Doubts that the rule will promote nutrition (e.g., lack of customer demand for healthy food means they won't buy it).

12. Support/Oppose: oppose
   Main Argument: Space Concerns
   Definition: Lack of space in store to adhere to depth-of-stock requirements.

13. Support/Oppose: oppose
   Main Argument: Perishability
   Definition: Concerns regarding how long items keep fresh (e.g., produce spoiling too quickly).

14. Support/Oppose: oppose
   Main Argument: Rule Ambiguity
   Definition: Claims that the SNAP retailer rule needs to be more clearly defined.

15. Support/Oppose: oppose
   Main Argument: Free Market
   Definition: Belief that the government is interfering with free market principles of supply and demand.

16. Support/Oppose: oppose
   Main Argument: Distribution and purchasing and infrastructure
   Definition: Lack of food distributors and purchasing infrastructure to order and stock staple food items.

17. Support/Oppose: oppose
   Main Argument: Equipment and Supplies
   Definition: Concerns about lack of adequate equipment and supplies, including coolers, refrigeration, and shelving.
\end{verbatim}

CLASSIFICATION TASK:

For each comment, you will read the ENTIRE comment and provide 4 classifications:

1. **support\_oppose**
   - Primary Goal: Match the comment's core message to a Definition in the rubric above
   - If a match is found: Assign the corresponding support\_oppose value from the matched category
   - Handling Multiple Matches: If a comment contains arguments aligning with multiple definitions, choose the definition that represents the PRIMARY or most emphasized point
   - Handling No Match: If the comment does not align with ANY of the 20 rubric definitions above, assign "Unknown"
   - Valid values: "support", "oppose", "facilitators to implementation", "Unknown"

2. **main\_argument**
   - If you found a rubric match: Copy the EXACT main\_argument text from the matched category above
   - If support\_oppose is "Unknown": Assign a 1-2 word summary of the comment's primary concern (e.g., "Cost", "Efficiency", "Clarity")

3. **submitter\_type** (from comment data/text)
   - Choose ONE category from this definitive list:
     * Retail/Business
     * Nonprofit
     * Private Citizen
     * Government
     * Medical Field
     * Unknown (if organization field is empty AND comment text provides no clear indication)
   - Use the organization field FIRST; if empty, analyze the comment text
   - Look for: company names, trade associations, business operations, nonprofit missions, government agencies, individual perspectives

4. **confidence** (your judgment of match quality)
   - Rate the quality of the match between the comment's argument and the chosen Definition from the rubric:
   - High: The comment's language and argument are highly similar to the rubric Definition. The comment is an unambiguous example of that argument.
   - Medium: There is a clear overlap or strong conceptual link, but the comment uses different wording or is somewhat similar
   - Low: There is minimal overlap; the comment is a very loose or inferential fit for the chosen Definition. Use sparingly—if the fit is truly poor, it should likely be "Unknown"

IMPORTANT NOTES:
- Read the ENTIRE comment before classifying
- "do not support" = Oppose (NOT Support)
- Focus on the PRIMARY argument if multiple are present
- Confidence measures the match quality with rubric definitions, not your certainty
\begin{verbatim}
You will receive multiple comments at once. Respond with a JSON array where each element has:
{{
  "comment_index": 0,
  "support_oppose": "value",
  "main_argument": "value",
  "submitter_type": "value",
  "confidence": "value"
}}
\end{verbatim}
Respond ONLY with valid JSON array, no other text.
\section{Evaluation metrics}
\label{app:metrics}
\paragraph{Classification accuracy.}
We measure stance accuracy (Support / Oppose) against our 150-comment human-labeled validation
set for both P1 and P2. We additionally measure theme accuracy for P2 only, computed as the
fraction of comments whose model-assigned theme exactly matches the human-labeled rubric category.
Theme accuracy is not evaluated for P1 because the number of model-generated themes varied
substantially across models.%

\paragraph{Structural metrics}
We apply two distributional metrics across all 1,260 classified comments to characterize how models organize thematic content and a similarity metric to measure consistency in vocabularies,.
 
\begin{itemize}
  \item \textbf{Shannon entropy} Measures information content; higher values indicate more evenly distributed themes; lower values indicate concentration in fewer categories.
  \begin{equation}
    H = -\sum_{i=1}^{n} p(t_i) \log_2 p(t_i)
  \end{equation}
  where $n$ is the number of distinct themes and $p(t_i)$ the proportion of comments assigned
  to theme $i$. Entropy ranges from 0 (all comments in one theme) to $\log_2(n)$ (uniform
  distribution).
 
  \item \textbf{Gini coefficient} ($G \in [0,1]$) measures inequality of comment counts across themes. $G = 0$ indicates equal distribution; $G = 1$ indicates complete concentration in a single theme.
  \begin{equation}
    G = \frac{\sum_{i=1}^{n} \sum_{j=1}^{n} |x_i - x_j|}{2n \sum_{i=1}^{n} x_i}
  \end{equation}
  where $x_i$ is the number of comments assigned to theme $i$.

    \item \textbf{Jaccard similarity} measures the overlap between two sets of themes $A$ and $B$, defined as the size of their intersection divided by the size of their union:
  \begin{equation}
    J(A, B) = \frac{|A \cap B|}{|A \cup B|}
  \end{equation}
  $J \in [0,1]$, where $J = 0$ indicates no shared themes and $J = 1$ indicates identical theme vocabularies. We compute $J$ across all 20 model-prompt combinations to assess whether theme divergence is driven primarily by model identity or prompt surface form.
  
\end{itemize}
\section{Sample Comment Classifications --- Open Prompts (P1)}
\label{app:sample-open}

We present a couple of comments from each disagreement category in the open-prompt evaluation set.
Because five prompt variants were used (Format, Lexical, Prompt Extraction Binary, Role,
and Verbosity), each table shows the theme assigned by each model under each prompt variant,
alongside the overall consensus model vote and stance.

\subsection*{Deep Theme Disagree}

\begin{table}[h]
\centering
\caption{Open-prompt example: \emph{Deep theme disagree} (Comment \texttt{FNS-2016-0018-0002}).
Four distinct themes emerge across models and prompt variants.}
\label{tab:open-deep}
\resizebox{\textwidth}{!}{%
\begin{tabular}{llllll}
\toprule
\textbf{Model} & \textbf{Format} & \textbf{Lexical} & \textbf{Prompt Extraction} & \textbf{Role} & \textbf{Verbosity} \\
\midrule
Gemini  & Reduced Access     & Reduced Food Access  & Food Access & Food Access        & Reduced Access    \\
GPT-5   & Access Loss        & Access Exception     & Food Access & Access Concerns    & Access Exception  \\
Llama   & Access Concerns    & Access               & Access      & Access Concerns    & Access            \\
Mistral & Food Access        & Food Access          & Access      & Accessibility      & Access Difficulty \\
\midrule
\multicolumn{6}{l}{\textit{Consensus vote: Reduced Access \quad Stance (all): Oppose \quad Disagreement type: Deep theme disagree}} \\
\bottomrule
\end{tabular}%
}
\end{table}

\subsection*{Stance Disagree}

\begin{table}[h]
\centering
\caption{Open-prompt example: \emph{Stance disagree} (Comment \texttt{FNS-2016-0018-0011}).
Models disagree on Support vs.\ Oppose: Gemini opposes while GPT-5, Llama, and Mistral support.}
\label{tab:open-stance}
\resizebox{\textwidth}{!}{%
\begin{tabular}{lllllll}
\toprule
\textbf{Model} & \textbf{Stance} & \textbf{Format} & \textbf{Lexical} & \textbf{Prompt Extraction} & \textbf{Role} & \textbf{Verbosity} \\
\midrule
Gemini  & Oppose  & Small Businesses       & Food Access Concerns  & Food Deserts         & Retailer Burden   & Retailer Burden    \\
GPT-5   & Support & Access Exception       & Access Concerns       & Access Concerns      & Access Concerns   & Access Concerns    \\
Llama   & Support & Access                 & Access                & Access               & Food Deserts      & Accessibility      \\
Mistral & Support & Food Access            & Food Access           & Access               & Access            & Healthy Foods      \\
\midrule
\multicolumn{7}{l}{\textit{Consensus vote: Retailer Burden \quad Disagreement type: Stance disagree}} \\
\bottomrule
\end{tabular}%
}
\end{table}

\section{Stage-2 Conditioning Template (Protocol B-4)}
\label{app:b4-template}

The Stage-2 prompt for the four LLMs in the 40-comment study (Section~\ref{sec:revision-protocol}) wraps the Stage-1 Verbosity prompt with a block listing all four Stage-1 labels (the model's own and the three other models'). The four labels are formatted as \texttt{model\_short:~label}, one per line, shuffled with a fixed seed, with the model's own line annotated \texttt{(your prior reading)}. Short model names (\texttt{GPT-5}, \texttt{Llama}, \texttt{Gemini}, \texttt{Mistral}) are used in place of full version strings. The full template is reproduced below.

\begin{verbatim}
COMMENT:
{comment}

OTHER READERS' CATEGORIZATIONS (4 single-label readings of this same
comment under the same prompt -- yours plus three others; format:
model: label):
{four_cells_text}

Classify this comment. You may agree with any of the readings above,
disagree, or pick something none of them said. Do not feel compelled
to converge with the other readers -- they may be wrong, partial, or
reflect different aspects than the one you find central. Whatever
label you pick, anchor it in a specific quote from the comment.

Output a single JSON object, no preamble, no markdown fences:

{
  "support_oppose": "support" | "oppose",
  "main_argument": "1-3 word label in your own words",
  "evidence_quote": "exact text from the comment that anchors your label",
  "agreement_with_others": "in one short sentence: does your label
       match any other reader's, or do you disagree?",
  "rationale": "one short sentence on why this is the main argument
       and not one of the alternatives the other readers proposed",
  "submitter_type": "Business" | "Nonprofit" | "Individual" |
       "Government" | "Unknown",
  "confidence": "High" | "Medium" | "Low"
}
\end{verbatim}

An example of the \texttt{four\_cells\_text} block, as it would be inserted into the template above for a single comment when the target model is Gemini:

\begin{verbatim}
Llama: Cost-Benefit Argument
GPT-5: Space Concerns
Gemini (your prior reading): Reduced Food Access
Mistral: Space Concerns
\end{verbatim}

The system prompt for both stages is identical and contains only a brief task description followed by the proposed-rule context text. All four LLMs were called at temperature 1.0; the only Stage-1/Stage-2 difference is the user prompt body shown above.

\subsection{Annotator Instructions}
\label{app:annotator-instructions}

The two instruction sheets below were the only task-relevant text the human annotator received before each stage. They are reproduced verbatim. The four LLMs were named only in Stage 2, after Stage-1 labels had been submitted.

\paragraph{Stage 1 (independent labeling, blind to all model output).}

\textit{Public-comment categorization (40 comments)}

\textit{What this is:} 40 public comments on a USDA SNAP retailer rule (FNS-2016-0018). For each one, give your read.

\textit{Your task, per comment:}
\begin{enumerate}
  \item \texttt{student\_label} --- One label, in your own words, for what this comment is about. Don't use a fixed list --- pick whatever phrasing feels right. Aim for a phrase, not a sentence (e.g., `Reduced food access', `SNAP retailer rule definitions', `Cost-benefit concerns').
  \item \texttt{student\_evidence\_quote} --- One short quote from the comment (1--2 sentences max) that anchors why you chose that label.
  \item \texttt{notes} --- Optional. Use this to flag anything confusing, ambiguous, or worth raising.
\end{enumerate}

\textit{Important:}
\begin{itemize}
  \item Don't look anything up about these comments or this rule while working through the file.
  \item Read each comment in full.
  \item If a comment is mostly procedural (e.g., a letterhead with little body), say so in notes --- your honest read is what we want.
\end{itemize}

\paragraph{Stage 2 (revision after exposure to the deduplicated LLM-label union).}

\textit{Public-comment categorization --- round 2 (40 comments)}

\textit{What this is:} Same 40 comments as the first batch. For each one, you'll see your original label and evidence quote, plus categorizations produced by four large language models (gpt-5.4, Llama-3.3, Gemini-3.1-pro, Mistral-medium) for the same comment. The LLMs were given the comment alone, then independently classified it.

\textit{What's in the LLM union column:}
\begin{itemize}
  \item Each bullet is a distinct concept the LLMs surfaced for this comment.
  \item In parentheses: which of the four models produced a label in that concept.
  \item In quotes: a short excerpt from the comment that one of the models cited as evidence for that label.
  \item Concepts are sorted by how many models surfaced them (most-supported first). A concept only surfaced by one model is still shown.
\end{itemize}

\textit{Your task, per comment:}
\begin{enumerate}
  \item Re-read the comment.
  \item Look at your original label (\texttt{your\_blind\_label}) and the LLM union side by side.
  \item Decide what to do --- fill the \texttt{decision} column with ONE of: Keep my label; Change to one shown above; Change to a different label.
  \item If you changed (either option 2 or 3), fill \texttt{final\_label} with your new label and \texttt{final\_evidence} with one short quote anchoring it.
  \item Fill the \texttt{justification} column --- one or two sentences on WHY you held or changed. This is the most analytically valuable part for us.
\end{enumerate}

\textit{Important:}
\begin{itemize}
  \item If you keep your label, justification should explain why the LLM concepts didn't change your reading.
  \item If you change, justification should explain what shifted you.
  \item Don't feel pressure to change just because multiple LLMs converged on something. Many model-converged concepts are still wrong or partial.
\end{itemize}

\section{Stratified-40 Sample Selection}
\label{app:stratified-40}

\subsection{Selection logic}
\label{app:strat40-logic}

The 40-comment subsample (Section~\ref{sec:revision-protocol}) was constructed by crossing the four-category disagreement taxonomy (Section~\ref{sec:taxonomy}) with two source-type strata (text in the comment field vs.\ text in an attached document), giving $4 \times 2 = 8$ cells of 5 comments each. For every comment in the corpus we computed the modal \texttt{main\_argument} per model across the 10 P2 runs, then assigned the comment to one of the four taxonomy categories: Consensus (all four models' modes identical), Binary split (two distinct modes), Deep theme disagree (three or four distinct modes), or Stance disagree (any disagreement on \texttt{support\_oppose}, taking precedence over thematic disagreement). Within each cell we sampled 5 comments uniformly at random with seed 42. All exclusions described in Sections~\ref{app:strat40-dedup}--\ref{app:strat40-exclude} were applied prior to sampling.

\subsection{Form-letter de-duplication}
\label{app:strat40-dedup}

Form-letter campaigns dominate the corpus: of 1{,}260 comments, 475 are near-duplicates of one of nine campaign templates. We identify duplicates by SHA-1 hashing the first 500 characters of each comment's effective text after normalization (lowercased, non-word characters replaced by spaces, whitespace collapsed); effective text is the attachment text if it exceeds 50 characters, otherwise the comment-field text. Comments yielding identical hashes are grouped, and within each group only the lowest-sorted comment ID is retained as the representative. The remaining 474 are removed from the sampling pool. Comments with normalized text shorter than 350 characters (70\% of the prefix length) are treated as unique and not subject to dedup, since short prefixes overhash.

\subsection{Exclusion sets}
\label{app:strat40-exclude}

Three classes of comment were excluded from the sampling pool prior to stratification, totaling 22 comment IDs.

\textbf{Scan-extraction failures (n=13).} The attachment was a scanned image; \texttt{pdfplumber} returned approximately 15 characters of OCR garbage with no recoverable substantive content. IDs: \texttt{FNS-2016-0018-0324}, \texttt{0445}, \texttt{0446}, \texttt{0447}, \texttt{0449}, \texttt{0451}, \texttt{0452}, \texttt{0455}, \texttt{0444}, \texttt{1254}, \texttt{1255}, \texttt{1258}, \texttt{1259}.

\textbf{Administratively-withheld attachments (n=8).} The attachment URL returned HTTP 403 across every probed file extension, consistent with administrative withholding rather than a transient fetch failure (all 8 are anonymous submissions clustered on three specific dates). IDs: \texttt{FNS-2016-0018-0213}, \texttt{0276}, \texttt{0278}, \texttt{0290}, \texttt{0522}, \texttt{0626}, \texttt{0769}, \texttt{0771}.

\textbf{Agency-redacted (n=1).} The substantive content was replaced with the marker ``\texttt{***Confidential Business Information Redacted***}''; nothing usable to label. ID: \texttt{FNS-2016-0018-1260}. This comment was originally drawn into the Stance-disagree / comment-only cell by the seed-42 sample; we substituted \texttt{FNS-2016-0018-0011} (the next draw from the same cell under the same seed).

\subsection{Final 40-comment list}
\label{app:strat40-list}

Table~\ref{tab:strat40-list} lists the 40 selected comment IDs by taxonomy cell and source-type.

\begin{table}[h]
\centering
\small
\caption{The 40 selected comment IDs (FNS-2016-0018-XXXX), 5 per cell across 4 taxonomy categories and 2 source types.}
\label{tab:strat40-list}
\begin{tabular}{lll}
\toprule
Taxonomy cell & Comment-only (5) & Attachment (5) \\
\midrule
Consensus            & 0231, 0059, 0691, 0635, 0571 & 0202, 0174, 1257, 0122, 0671 \\
Binary split         & 0025, 0021, 0103, 0303, 0362 & 1156, 0085, 0228, 0803, 0249 \\
Deep theme disagree  & 0495, 0664, 0163, 0944, 1090 & 0054, 0136, 0461, 0351, 0261 \\
Stance disagree      & 0124, 0242, 0553, 0101, 0011 & 0167, 0355, 0175, 0196, 0356 \\
\bottomrule
\end{tabular}
\end{table}

\section{Theme Clustering}
\label{app:clustering}

\subsection{Embedding pipeline}
\label{app:cluster-pipeline}

To compute the hold/absorb/novel outcome (Section~\ref{sec:revision-protocol}), Stage-1 and Stage-2 \texttt{main\_argument} strings from all four LLMs and the human annotator are clustered jointly so that paraphrases (e.g., ``Reduced food access'' and ``Reduces food access'') count as the same concept. Each unique label string is lowercased and embedded with \texttt{sentence-transformers/all-MiniLM-L6-v2} \citep{reimers2019sentence} with L2-normalized output, then clustered with \texttt{sklearn.cluster.AgglomerativeClustering} using cosine distance and average linkage. The cluster-count parameter is left unset; the algorithm cuts the dendrogram at a fixed distance threshold of 0.4 (cosine similarity $\ge 0.6$). Threshold 0.4 was selected to merge surface paraphrases without collapsing semantically distinct concerns; sensitivity to this choice is reported below.

\subsection{Threshold sensitivity}
\label{app:cluster-sensitivity}

Table~\ref{tab:cluster-sensitivity} reports cluster counts and per-model hold/absorb/novel rates at the chosen threshold (0.4) and at $\pm 0.1$ on either side. Tighter thresholds (0.3) leave more first-stage labels in singleton clusters, raising the novel rate; looser thresholds (0.5) merge more aggressively, raising the absorb rate.

\begin{table}[h]
\centering
\small
\caption{Hold / absorb / novel outcome rates per model at three cluster-distance thresholds. Each cell counts (out of 40) Stage-2 labels whose threshold-T cluster matches the model's own Stage-1 cluster (hold), another model's Stage-1 cluster only (absorb), or neither (novel).}
\label{tab:cluster-sensitivity}
\begin{tabular}{llccc}
\toprule
Threshold & Model & Hold (\%) & Absorb (\%) & Novel (\%) \\
\midrule
\multirow{4}{*}{0.3 (66 clusters)}
 & GPT     & 60.0 & 30.0 & 10.0 \\
 & Llama   & 32.5 & 12.5 & 55.0 \\
 & Gemini  & 75.0 & 25.0 &  0.0 \\
 & Mistral & 45.0 & 32.5 & 22.5 \\
\midrule
\multirow{4}{*}{\textbf{0.4 (52 clusters)}}
 & GPT     & 67.5 & 25.0 &  7.5 \\
 & Llama   & 40.0 & 10.0 & 50.0 \\
 & Gemini  & 77.5 & 22.5 &  0.0 \\
 & Mistral & 52.5 & 27.5 & 20.0 \\
\midrule
\multirow{4}{*}{0.5 (36 clusters)}
 & GPT     & 82.5 & 15.0 &  2.5 \\
 & Llama   & 42.5 & 15.0 & 42.5 \\
 & Gemini  & 77.5 & 22.5 &  0.0 \\
 & Mistral & 55.0 & 27.5 & 17.5 \\
\bottomrule
\end{tabular}
\end{table}

\subsection{Manual merge passes}
\label{app:cluster-merges}

The threshold-0.4 pass produces 52 clusters from 150 unique label strings. Inspection of the cluster contents reveals a small number of cross-cluster fragmentations the embedder did not catch: closely related concepts that share little surface vocabulary land in different clusters. We applied two rounds of manual merges, reducing the 52 clusters to 39 after round 1 and to 34 after round 2. Each merge rule names a target cluster and unions a list of source threshold-0.4 cluster IDs together with any singleton labels that belong with them; a separate polarity-split rule splits one polysemous cluster (the original ``Food access'' cluster, which mixed support-framed ``healthy food access'' with oppose-framed ``reduced food access'') into two unambiguously-framed clusters. Five representative merge rules:

\begin{itemize}
  \item \textbf{Convenience store exclusion} unions threshold-0.4 clusters 16 and 24 with the singletons ``Exclusion Concerns'' and ``Forces businesses out''.
  \item \textbf{Burden / unworkable requirements} unions clusters 10, 22, 46 with ``Stocking requirements burden'', ``Unworkable operational burdens'', and ``Unfair Burden''.
  \item \textbf{Reduced customer access} unions clusters 2 and 4 with ``Negative customer impact'' and ``customer inconvenience''.
  \item \textbf{Variety definition} unions clusters 18 and 1 (an inventory/variety pair) with ``increased variety''.
  \item \textbf{Healthy food access} (round-1 polarity split): all labels in cluster 25 matching the surface form ``Healthy/healthier food access'' plus the ``Fresh Foods'', ``Healthy Food'', and ``Healthy Food Options'' singletons; the remaining cluster-25 labels using ``Reduced/Reduces food access'' move to a separate \textbf{Reduced food access} cluster, and any cluster-25 labels that are framing-ambiguous remain in a residual \textbf{Food access (general)} cluster.
\end{itemize}

\end{document}